**RESEARCH ARTICLE**                                                                    **Open Access**

# Neural sentence embedding models for semantic similarity estimation in the biomedical domain

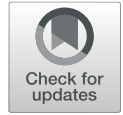

Kathrin Blagec, Hong Xu, Asan Agibetov and Matthias Samwald*

## Abstract

**Background:** Neural network based embedding models are receiving significant attention in the field of natural language processing due to their capability to effectively capture semantic information representing words, sentences or even larger text elements in low-dimensional vector space. While current state-of-the-art models for assessing the semantic similarity of textual statements from biomedical publications depend on the availability of laboriously curated ontologies, unsupervised neural embedding models only require large text corpora as input and do not need manual curation. In this study, we investigated the efficacy of current state-of-the-art neural sentence embedding models for semantic similarity estimation of sentences from biomedical literature. We trained different neural embedding models on 1.7 million articles from the PubMed Open Access dataset, and evaluated them based on a biomedical benchmark set containing 100 sentence pairs annotated by human experts and a smaller contradiction subset derived from the original benchmark set.

**Results:** Experimental results showed that, with a Pearson correlation of 0.819, our best unsupervised model based on the Paragraph Vector Distributed Memory algorithm outperforms previous state-of-the-art results achieved on the BIOSSES biomedical benchmark set. Moreover, our proposed supervised model that combines different string-based similarity metrics with a neural embedding model surpasses previous ontology-dependent supervised state-of-the-art approaches in terms of Pearson's r ($r = 0.871$) on the biomedical benchmark set. In contrast to the promising results for the original benchmark, we found our best models' performance on the smaller contradiction subset to be poor.

**Conclusions:** In this study, we have highlighted the value of neural network-based models for semantic similarity estimation in the biomedical domain by showing that they can keep up with and even surpass previous state-of-the-art approaches for semantic similarity estimation that depend on the availability of laboriously curated ontologies, when evaluated on a biomedical benchmark set. Capturing contradictions and negations in biomedical sentences, however, emerged as an essential area for further work.

## Background

The tremendous and unprecedented amount of literature published in the biomedical domain each year has the potential to drive scientific progress by facilitating instant and direct access to the latest scientific findings and data. This accelerating influx of information, however, also leaves researchers with considerable challenges of how to cope with these large amounts of data, calling for tools that enable efficient retrieval, filtering and summarization.

An essential component for the effective processing of textual information is the ability to determine semantic similarity, i.e., the likeness or contradiction in meaning, of two textual components (e.g., words, sentences, paragraphs) beyond their syntactic or lexical similarity. While humans can intuitively assess how similar, for example, two words are in their meaning, delegating this task to machines requires the availability of effective metrics that are able to accurately capture and quantify their semantic relatedness. Such semantic measures can be broadly divided into two categories: distributional and knowledge-based metrics, depending on whether they use corpora of texts or ontologies as proxies, respectively. While knowledge-based measures have previously been shown to be more effective for semantic similarity estimation in the biomedical field, they are dependent on the availability of domain-specific ontologies, whose creation – despite the emergence of automatic and semi-automatic ontology learning – still remains a tedious, work-intensive and error-prone task [1] .

* Correspondence: matthias.samwald@meduniwien.ac.at
Section for Artificial Intelligence and Decision Support, Center for Medical Statistics, Informatics, and Intelligent Systems, Medical University of Vienna, Währinger Straße 25a, 1090 Vienna, Austria

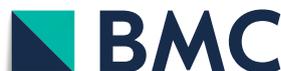

Furthermore, while previous studies suggest that combining different measures in a supervised model perform well in semantic similarity estimation in the general domain, Sogancioglu et al. have recently demonstrated the insufficiency of these general domain state-of-the-art sentence similarity computation systems when applied to similarity estimation tasks in the biomedical field (Sogancioglu et al. 2017; Panchenko and Morozova 2012). By merging different knowledge-based (ontology-based) and distributional (corpus-based) semantic similarity measures that are specifically tailored to the biomedical domain in a supervised model, they demonstrated the superiority of these domain-specific approaches compared to general domain state-of-the art systems. Besides string-based measures and ontologies, they also utilized a distributional vector representation methodology, i.e. the Paragraph Vector model [2]. Boosted by advances in hardware technology that allow fast processing of large amounts of text data, such neural network-based methods for embedding words, sentences or even larger text elements in low-dimensional vector space have recently caught attention for their ability to effectively capture semantic information [3, 4].

Numerous neural network architectures for generating these embeddings have been published in recent years [5–8]. In contrast to current state-of-the-art models for assessing semantic similarity that depend on the availability of labor-intensively curated ontologies, neural embedding models only require large, unstructured text corpora as input. This makes them especially promising tools for further use in text-related applications in the biomedical domain such as question-answering systems.

In this study, we investigate the usefulness of current state-of-the-art neural sentence embedding models for semantic similarity estimation in the biomedical domain.

We evaluate and compare these models based on a biomedical benchmark dataset published previously by Sogancioglu et al. [9], thereby building on and extending their work. We demonstrate that these models, both when used as standalone methods or when combined with string-based measures, can keep up with and even surpass previous approaches that utilized sophisticated, manually curated ontologies in terms of their capability to effectively capture semantic information. Furthermore, we present preliminary results on the performance of our best neural embedding models on a small contradiction subset derived from the original benchmark dataset.

## Results
### BIOSSES benchmark set

For each of the unsupervised approaches (i.e. sent2vec, skip-thoughts, Paragraph Vector, fastText continuous bag-of-words / C-BOW and fastText skip-gram) we evaluated several different models trained with different parameters on the BIOSSES benchmark set by assessing the correlation between estimated similarity scores and scores assigned by human annotators. Furthermore, we calculated simple string-based similarity scores using Jaccard and q-gram distance for later use in our supervised model. Pearson correlation coefficients for the best results obtained with each method are presented in Table 1. We also calculated Spearman's r but found it to yield similar values, therefore we only report Pearson's r.

Table 2 shows results achieved by current state-of-the-art approaches that we used as baselines for comparing the results of our models. For our unsupervised models, the best result ($r = 0.819$) was attained by a Paragraph Vector Distributed Memory (PV-DM) model with embedding dimension of 100, trained on a filtered version of our PMC Open Access corpus which contained only lines with less than 200 characters. Our best sent2vec mode, which is characterized by significantly lower computational complexity

**Table 1** Highest correlation coefficients obtained with different methods

| Method | r |
|---|---|
| *String-based methods* | |
| Jaccard | 0.751 |
| Q-gram (q = 3) | 0.723 |
| *Unsupervised* | |
| fastText (skip-gram, max pooling) | 0.766 |
| fastText (CBOW, max pooling) | 0.253 |
| Sent2vec | 0.798 |
| Skip-thoughts | 0.485 |
| Paragraph vector (PV-DM) | 0.819 |
| Paragraph vector (PV-DBOW) | 0.804 |
| *Unsupervised combination of several methods (mean)* | |
| Jaccard, q-gram, Paragraph vector (PV-DBOW) and sent2vec | 0.846 |
| *Supervised combination of several methods* | |
| Supervised linear regression (Combination of Jaccard, Q-gram, sent2vec, Paragraph vector DM, skip-thoughts, fastText) | 0.871 |

*r* Pearson correlation, *CBOW* Continuous Bag of Words, *PV-DM* Paragraph Vector Distributed Memory, *PV-DBOW* Paragraph Vector Distributed Bag of Words

Table 2 Baseline values for our analysis, as reported by Sogancioglu et al. [9]

| Method | r |
|---|---|
| Jaccard | 0.710 |
| Q-gram | 0.754 |
| Paragraph Vector (PV-DBOW) | 0.787 |
| Supervised linear regression (Combined ontology method, Paragraph vector, Q-gram) | 0.836 |

PV-DBOW Paragraph Vector Distributed Bag of Words

compared to the Paragraph Vector model, yielded a correlation of $r = 0.798$. Detailed hyper-parameter settings of all models are shown in Additional file 1: Tables S1-S4. For comparison, we also evaluated a publicly available sent2vec model (700 dimension) trained on general domain text (i.e. English Wikipedia content) and found it to yield a substantially lower correlation score ($r = 0.44$).

In general, we found that the following post-processing steps empirically proved to enhance the quality of our sentence embeddings across all models (1) Filtering the corpus for excessively long lines and (2) Separation of compound words that are connected by hyphens.

We furthermore present the results of a supervised models that combine several of the unsupervised models in a hybrid approach, yielding a correlation of $r = 0.871$ (see Table 1).

Distribution of similarity scores for each best model and correlations between different models are shown in Additional file 1: Figures S1 and S2.

As can be seen from Tables 1 and 2, both our best unsupervised and our best supervised model outperform previous state-of-the art results. It has to be noted that both supervised models are hybrids of string-based similarity measures and corpus-based embedding models, but do not make use of any ontologies as opposed to the model reported by Sogancioglu et al. [9] that served as a baseline for our analyses (Fig. 1).

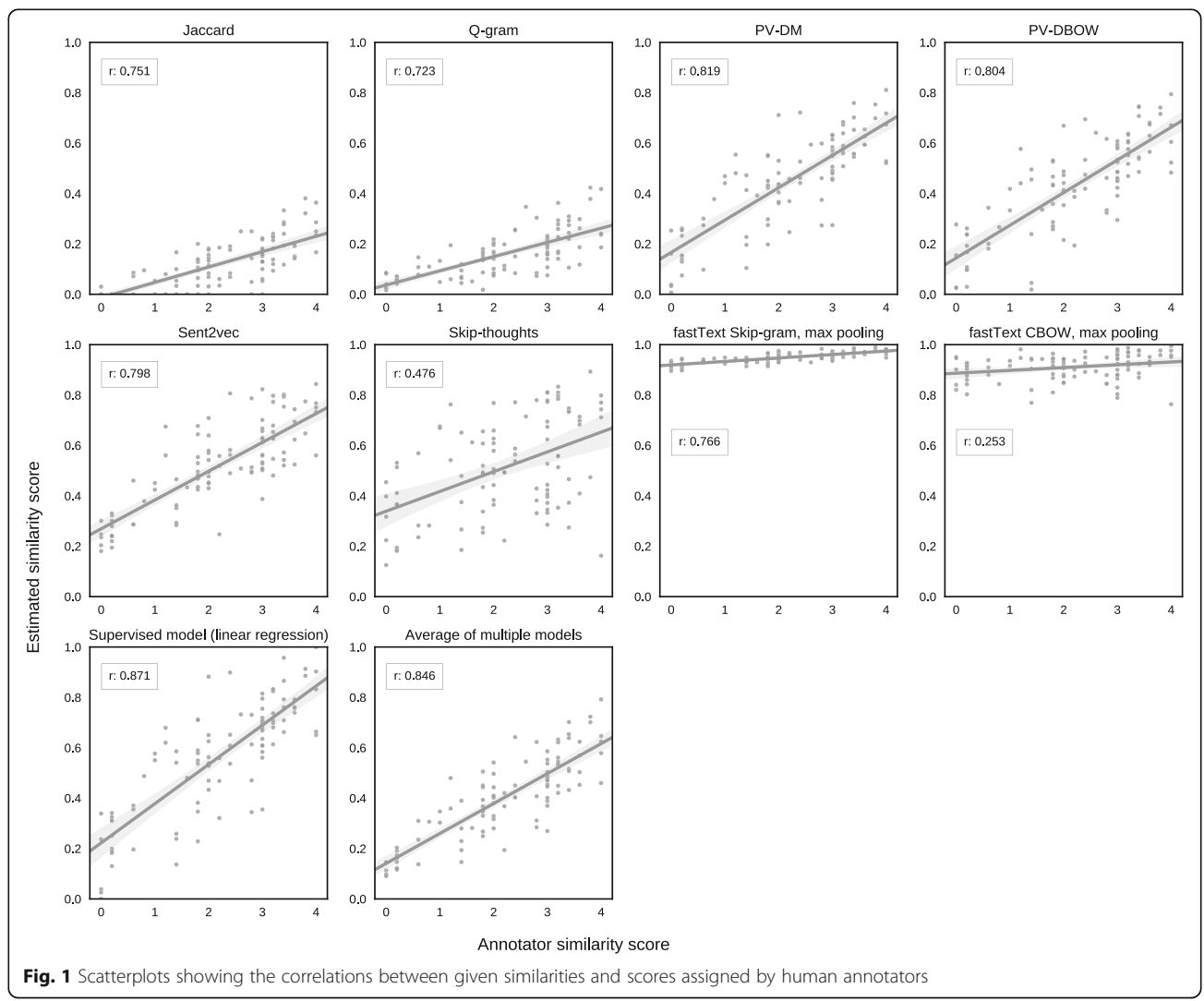

Fig. 1 Scatterplots showing the correlations between given similarities and scores assigned by human annotators

### Contradiction subset

Besides evaluating our models on the BIOSSES benchmark set, we experimented with a small contradiction dataset that was manually created based on a small subset of the original benchmark set and present preliminary results on these experiments. Table 3 shows mean cosine similarities obtained for the negation and antonym subsets and the subset of sentences that were rated as highly similar by the human experts (score of 3.5 or higher). Lower values indicate lower estimated semantic similarity, whereas higher values indicate higher estimated semantic similarities. We would therefore expect the negation and antonym subsets to show lower estimated cosine similarities than the subset of highly similar sentences. Interestingly, for all considered models, average cosine similarities obtained for the negation and antonym subsets were higher than the average cosine similarities obtained for the subset of semantically highly similar sentences. We refrained from testing for statistical significance due to the limited power of such a test given the low sample size.

### Discussion

In recent years, neural network-based approaches have gained attraction in the field of natural language processing due to their ability to effectively capture semantic information by representing words, sentences or even larger text elements in low-dimensional vector space when trained on large text corpora. As these models are easy to train, involving minimal manual work and only requiring large amounts of unstructured text data which is abundantly available given the large number of scientific articles published each year, we were interested in how they would perform compared to current state-of-the-art semantic similarity estimation systems that require the availability of elaborate ontologies.

We evaluated several different neural network-based embedding models and compared our results to previous state-of-the-art results on the BIOSSES biomedical benchmark set reported by Sogancioglu et al. [9]. Experimental results showed that our proposed supervised model that combines different string-based similarity metrics with neural embedding models surpasses previous ontology-dependent supervised state-of-the-art approaches. Furthermore, our best unsupervised model based on the Paragraph Vector Distributed Memory algorithm outperforms previous state-of-the-art results achieved by a standalone neural-based embedding model on the BIOSSES biomedical benchmark set.

We found our results for the string-based measures, i.e. q-gram and Jaccard, to be slightly different from those reported by Sogancioglu et al. This could be explained by variations in sentence pre-processing since we used the stop word list from the Python NLTK library which covers slightly less words (153 vs. 174, respectively) than the RANKS list utilized by Sogancioglu et al.[1, 2].

The higher correlation for our Paragraph Vector PV-DBOW model as compared to Sogancioglu et al. may be explained by (1) differences in the choice of model hyper-parameters, and (2) differences in the training corpus and corpus pre-processing. We trained our PV-DBOW model using the same vector dimension and nearly identical hyper-parameters as Sogancioglu et al., except for setting the number of minimal word occurrences to five instead of one to accelerate training. Furthermore, we trained on a much larger corpus despite filtering for line lengths since we used the complete PMC dataset instead of only a subset.

We used the PubMed Central (PMC) Open Access dataset to train our models, applying a simple heuristic to remove excessively long lines from our corpus that did not contain proper sentences, but were caused by table content instead. Furthermore, we experimented with simple post-processing steps, such as separating compound words that are connected by hyphens, which we found to markedly improve embedding quality. More advanced and exhaustive filtering and cleaning of the data set may improve the corpus quality further and consequently lead to even better training results. This is in line with previous findings by Chiu et al. who evaluated and compared embeddings trained on three different biomedical corpora: (1) the OMC Open Access dataset, (2) the PubMed dataset which contains the abstracts of more than 25 million biomedical scientific publications, and (3) a combination of both corpora. They found that using a combination of both corpora worsened the quality of their embeddings, which they speculatively attributed to the amount of non-prose text in the PMC dataset [10].

**Table 3** Average estimated cosine similarities for sentence pairs included in the negation and antonym subset and a reference set of highly similar sentences per model. Lower values indicate lower estimated semantic similarity; higher values indicate higher estimated semantic similarities

|  | Sent2vec | Skip-thoughts | PV-DM | PV-DBOW | fastText CBOW | fastText skip-gram |
| --- | --- | --- | --- | --- | --- | --- |
| Subset of highly similar sentences ($n = 11$) | 0.706 | 0.899 | 0.652 | 0.568 | 0.938 | 0.971 |
| Negation subset ($n = 13$) | 0.967 | 0.999 | 0.930 | 0.936 | 0.945 | 0.979 |
| Antonym subset ($n = 7$) | 0.983 | 0.999 | 0.968 | 0.960 | 0.976 | 0.989 |

*PV-DM* Paragraph Vector Distributed Memory, *PV-DBOW* Paragraph Vector Distributed Bag of Words, *CBOW* Continuous Bag of Words

For sent2vec and fastText, we experimented with different hyper-parameter settings based on previously published findings on training embeddings in the biomedical domain [10]. Spending additional time on hyper-parameter tuning might yield even better embeddings.

The slightly better results achieved with both PV-DM and PV-DBOW model compared to our best sent2vec model have to be interpreted bearing in mind the differences in computational complexity of these two models, which is $O(1)$ for sent2vec and $O(n * |V|)$ for Paragraph Vector, where n is the number of 'paragraphs' (in our case sentences) and $|V|$ is the size of the input vocabulary [2, 5], resulting in a considerable smaller training time for sent2vec models.

Surprisingly, our best fastText skip-gram and CBOW models yielded strikingly different correlations on the benchmark set. This inconsistency may be explained by the mechanism by which skip-gram and CBOW models are trained (i.e. learning to predict the context using the current word vs. learning to predict the current word using the context, respectively), potentially making skip-gram models more capable of dealing with rare words. Differences between the performance of skip-gram and CBOW models in tasks related to semantic similarity have been reported previously, and might be aggravated by the high frequency of rare words commonly found in biomedical texts, such as protein and gene names, or terms to describe metabolic pathways [11]. For example, considering the sentence 'Oncogenic KRAS mutations are common in cancer.': While the term KRAS (i.e., an oncogene) may be sufficient to estimate the context (i.e., a context most likely related to cancer, biomedical pathways or similar), reversing the task and predicting the term KRAS from the surrounding words, appears much more difficult, considering that there is a wide range of different oncogenes that may fit into the context.

Two different approaches for assessing the quality of embeddings can be distinguished: (1) Extrinsic evaluation, which refers to evaluation in the context of a specific natural language processing task, and (2) intrinsic evaluation, which attempts to assess overall embedding quality beyond a specific downstream task by applying similarity or distance metrics. We evaluated our sentence embeddings intrinsically based on a publicly available domain-specific benchmark set and used the degree to which estimated semantic similarity corresponds to similarity scores assigned by human experts in terms of Pearson correlation as our evaluation metric. Albeit this procedure is in line with established practices for assessing embedding quality, such intrinsic evaluation tasks have recently been shown to be insufficient for prediction of performance in downstream tasks [10, 12].

Further extrinsic evaluation of neural sentence embeddings, for example, as a part of biomedical question-answering systems, is therefore an essential next step in confirming their value for domain-specific downstream tasks.

We furthermore conducted a preliminary evaluation of the neural models' ability to detect contradictions between sentence pairs on biomedical topics and found that good performance on semantic similarity estimation does not necessarily imply the ability to distinguish contradicting sentence pairs from highly similar sentence pairs. For sentence pairs that are contradicting each other due to the presence of antonyms, these results can be traced to the context-based learning mode of common neural embedding models, where context is treated as "bag of words", and words with similar contexts are close together in vector space. This potential shortcoming is already known from the domain of sentiment analysis. A commonly applied approach to overcome this issue in the field of sentiment analysis is to use supervised training based on sentiment polarity labels (e.g., positive and negative) [13–15] that can, e.g., be derived from sentiment lexicons.

For contradictions due to the presence of a negation, a simple heuristic approach for detecting negations may improve performance. Negation detection in sentences with a complicated syntactic structure may, however, require more sophisticated methods [16].

Since our preliminary evaluation of the usefulness of neural embeddings for contradiction detection in biomedical sentences was limited to a small set of sentences pairs, results have to be interpreted with caution. To be able to thoroughly investigate the ability of novel embedding models to detect contradictions in biomedical statements, an extensive benchmark covering common subtypes of contradictions, such as negation, numerical mismatch or antonym-based contradictions, may be of value.

## Conclusions

In this study, we have highlighted the value of neural network-based models for semantic similarity estimation in the biomedical domain by showing that they can keep up with and even surpass previous state-of-the-art approaches for semantic similarity estimation that required the availability of labor-intensive manually curated ontologies. However, we identified current standard neural sentence embedding models' ability to detect contradictions in biomedical sentences as an important area for further research.

## Methods

### Training corpus and pre- and post-processing procedures
#### PubMed central open access dataset
We used the complete PubMed Central (PMC) open access subset as of November 2017 as our training corpus.

This dataset contains over 1.7 million biomedical articles that have a Creative Common or similar license. Data are available in both plaintext and XML format via the NCBI FTP site[3]. Corpus statistics are shown in Table 4.

### Corpus pre- and post-processing

Preparing and pre-processing of the training corpus are outlined in Fig. 2 and comprised the following steps: (1) Generating a single text file that contains all articles of the PMC open access dataset, one sentence per line. (2) Tokenization. The Stanford CoreNLP library was used for sentence boundary detection and tokenization [17].

Initially, we used the resulting, unfiltered corpus for training our embeddings, which, however, resulted in a disproportionately long training time caused by the occurrence of exceptionally long lines. Inspection of the files revealed that they included content beside sentences, such as large numeric tables. We therefore decided to proceed with the analysis using post-processed versions of the original corpus that were filtered to exclude excessively long lines, to accelerate the training time. Furthermore, we post-processed our training corpus to separate compound words that are connected by hyphens, as we empirically found this processing step to improve our results. For the string-based methods, we removed stop words as defined by the NLTK python library[4] and the following punctuation marks: full stop, comma, colon, semicolon, question mark, exclamation mark, slash, dash.

### Embedding models
#### fastText

fastText is a library that allows unsupervised learning numerical word representations using either the skip-gram or continuous bag-of-words (C-BOW) model [18]. The C-BOW model learns to predict a target word based on a defined number of surrounding context words (i.e. the windows size) [6].

The skip-gram model learns to predict the context (i.e. the neighbouring words) using the current word. In both cases, the amount of words in of the context is defined by the window size parameter. fastText can also be seen as an extension of the word2vec model. While for word2vec, the defining entity is an entire word, fastText additionally allows for representing each word as a composition of character n-grams with the numerical representation of a word being a sum of these n-grams. Furthermore, it allows for capturing word n-grams. These added features make fastText more amenable to represent rare lexical variations of words, as well as phrases.

We trained both skip-gram and C-BOW word vectors on our corpus using different hyper-parameter settings shown in Additional file 1: Table S3 using a grid search approach. Sentence vectors were then derived by max, min, sum or average pooling across vectors of all words that constitute the respective sentence.

#### Sent2Vec

Sent2Vec is an unsupervised model for learning general-purpose sentence embeddings. It can be seen as an extension of the C-BOW model that allows to train and infer numerical representations of whole sentences instead of single words. The algorithm is characterized by its low computational complexity, while simultaneously showing good performance on a wide range of evaluation tasks [5].

We trained sent2vec embeddings on the PMC Open Access corpus using a grid search approach for hyper-parameter tuning. We experimented with different values of the following hyper-parameters: dimension, wordNgrams, epoch, minCount, dropoutK and sampling threshold. Values investigated are listed in Additional file 1: Table S2. Due to limited computational resources, we did not train all our models for 9 epochs.

Furthermore, to put results obtained with models trained on a biomedical corpus into context, we also evaluated the performance of a publicly available sent2vec model that was pre-trained on general domain text (i.e. English Wikipedia content)[5].

#### Skip-thoughts

Skip-thoughts is a RNN (recurrent neural network)-based model for unsupervised learning of a generic, distributed sentence encoder [7]. Requiring contiguous sentences $S_{i-1}$, $S_i$, $S_{i+1}$ as input, the model is trained to predict the previous ($S_{i-1}$) and subsequent sentence ($S_{i+1}$) based on the current sentence ($S_i$).

We used a TensorFlow implementation of the skip-thoughts model available at GitHub and trained 2400-dimensional skip-thought vectors using a unidirectional mode ('uni-skip') on our post-processed training corpus[6]. We also experimented with vocabulary expansion using GoogleNews word2vec vectors[7]. Hyper-parameters of our skip-thoughts model can be found in Additional file 1: Table S4.

**Table 4** Characteristics of the PMC Open Access dataset

| | |
|---|---|
| File size | 45 GB |
| Number of articles | > 1,700,000 |
| Total number of tokens | 8,126,457,106 |
| Number of unique words | 31,974,798 |
| Number of sentences | 277,809,416 |
| Average line length before post-processing (number of characters) | 162 |
| Longest line length before post-processing (number of characters) | 111,562 |

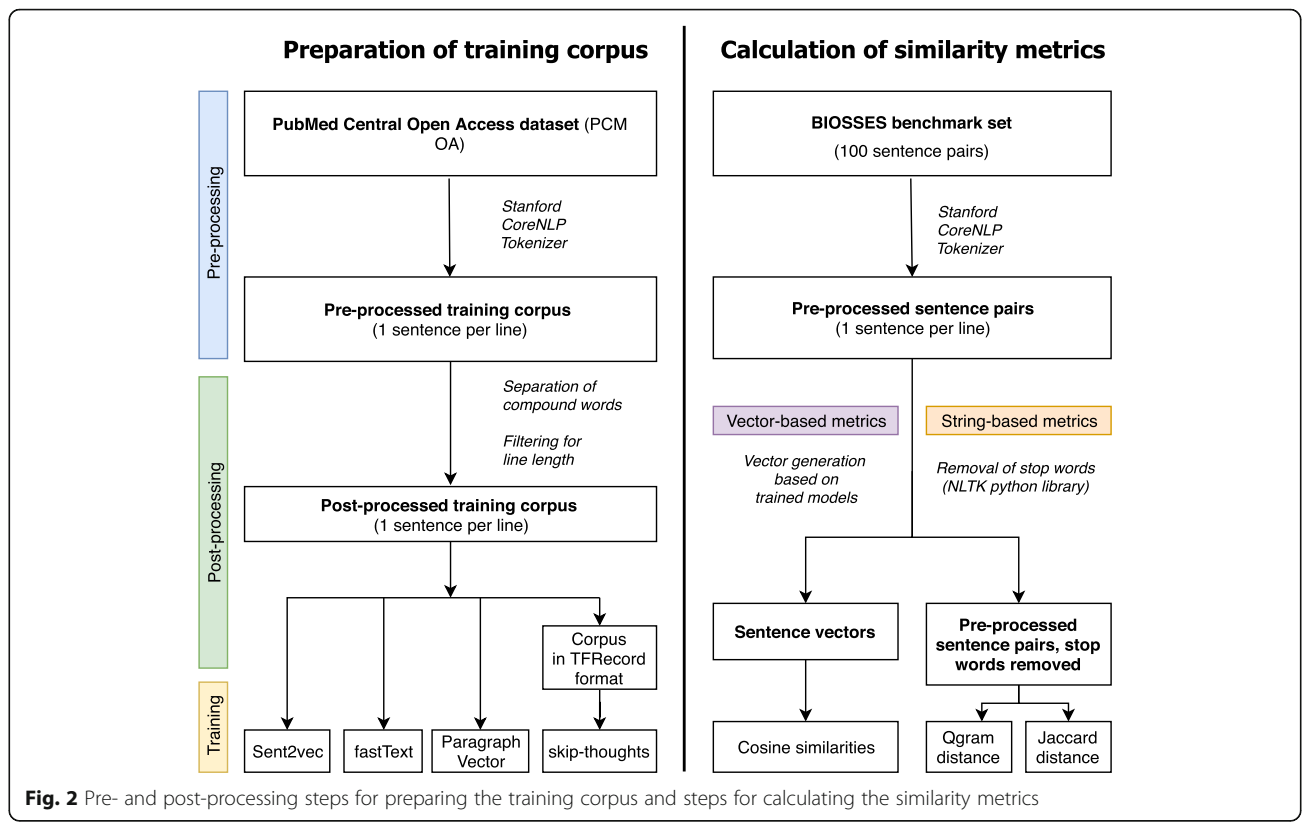

Fig. 2 Pre- and post-processing steps for preparing the training corpus and steps for calculating the similarity metrics

*Paragraph vector*

Paragraph Vector is an unsupervised model for representing variable-length texts, such as sentences, paragraphs or larger documents as fixed-size vectors [2]. The authors proposed two different versions of the model: (1) A Distributed Memory Model (PV-DM), where, given several contiguous words from a paragraph, the model is trained to predict the following word based on the concatenation of the vectors of the previous words with the vector of the given paragraph. Thus, and as opposed to classical bag-of-words models, information on the ordering of words does not get lost in the PV-DM model. (2) A Distributed Bag of Words (PV-DBOW) version, which is trained to predict words randomly sampled from the given paragraph.

We used a Python implementation of the Paragraph vector model available at GitHub[8] and trained 100-dimensional vectors on our corpus using both the PV-DM and PV-DBOW versions of the Paragraph Vector model. Due to the large computing time of these models, we trained only one PV-DM and one PV-DBOW model. Hyper-parameters of these models are shown in Additional file 1: Table S1.

**Evaluation of sentence embeddings**
*BIOSSES benchmark set*
To evaluate our sentence embeddings, we used the BIOSSES dataset, a benchmark set for biomedical sentence similarity estimation [9]. This dataset comprises 100 sentence pairs stemming from the biomedical domain that were selected by the authors from the TAC (Text Analysis Conference) Biomedical Summarization Track Training Dataset.

Each of the 100 sentence pairs was evaluated by five human experts who judged their semantic similarity using an ordinal scale ranging from 0 (no relation) to 4 (sentences are semantically equivalent). For each sentence pair, individual ratings were then averaged across all experts, resulting in a continuous value score ranging from 0 to 4. Table 5 shows two pairs of example sentences together with their suggested similarity scores.

*Contradiction subset*
The original BIOSSES benchmark set was compiled from citing sentences, i.e. sentences that either reference the same or a different research article, to ensure that different degrees of semantic similarity are represented in the pool of selected sentence pairs. Since this compilation method tends to only yield sentence pairs that are either semantically similar to some degree or unrelated, the resulting set is not suitable to evaluate the ability to accurately capture contradictions between sentence pairs. Contradiction refers to a semantic relation where two statements, e.g., two sentences, are opposed to one another. Subtypes of contradiction include (1) contradiction

**Table 5** Example sentences from the BIOSSES benchmark set

| Sentence 1 | Sentence 2 | Comment | Score |
| --- | --- | --- | --- |
| This article discusses the current data on using anti-HER2 therapies to treat CNS metastasis as well as the newer anti-HER2 agents | Breast cancers with HER2 amplification have a higher risk of CNS metastasis and poorer prognosis. | The two sentences are not equivalent, but share some details | 2 |
| The up-regulation of miR-146a was also detected in cervical cancer tissues. | The expression of miR-146a has been found to be up-regulated in cervical cancer. | The two sentences are completely or mostly equivalent. | 4 |

by negation, (2) antonym-based contradiction and (3) numeric mismatch.

Since we were interested in getting an insight into our models' performance regarding contradiction detection, we manually created two small contradiction subsets using sentences from the original BIOSSES dataset for experimental purposes: (1) A negation subset containing 13 sentence pairs, where sentence 1 is the original sentence, and sentence 2 its negation, (2) an antonym subset containing 7 sentence pairs, where keywords of the original sentences where replaced by one of their antonyms to shift the sentences' meaning to the opposite. For creating the subsets, candidate sentences with a simple syntactic structure were selected manually. Example sentence pairs for both subsets are listed in Tables 6 and 7. We calculated average cosine similarities obtained for the negation and antonym subsets with cosine similarities obtained for a subset of sentences from the BIOSSES dataset that were rated as highly similar by the human experts (score of 3.5 or higher). We refrained from testing for statistical significance due to the limited power of such a test given the low sample size.

## Assessing sentence vector similarity and evaluation with benchmark data set

**Evaluation of different embedding models** Using the embedding techniques described above, we generated sentence vectors for each of the 100 biomedical sentences pairs included in the BIOSSES dataset. For each of these sentence pairs, we calculated cosine similarity between vectors generated with the respective embedding model. To evaluate the different embedding models' ability to capture useful semantic properties of sentences, we measured the correlation between calculated semantic similarities and the averaged similarity assessments assigned by the five human experts using both Pearson correlation coefficient ($r$) and Spearman's correlation coefficient ($r_s$). Since we found the differences between Pearson's and Spearman's r to be negligible, we only report Pearson's r.

**Unsupervised hybrid model** Furthermore, we created a simple unsupervised hybrid model by averaging over the estimated cosine similarities obtained with the string-based and neural network-based models.

**Supervised model** In addition to the unsupervised models, we trained a supervised model using linear regression to predict the averaged annotation scores. We used a combination of string-based similarity coefficients together with vector-based similarity metrics obtained with our embedding models as features for training this model.

A linear regression can be described by the formula

$$y = \sum_{j=1}^{k} b_0 + b_j x_j$$

where y denotes the predicted variable, $b_0$ the intercept term, $b_j$ the estimated coefficient and $x_j$ the explanatory variable. In our case, $y$ corresponds to the predicted sentence similarity score, which is predicted based on the similarity scores obtained by the string- and our best neural network-based models (i.e., sent2vec, PV-DM, skip-thoughts, fastText).

Similarity measures that we used in our analysis are described in detail below.

**Similarity measures** String-based

Jaccard index

The Jaccard index is a similarity coefficient that measures similarity between sets by comparing which members of the sets are shared and which are distinct. It is computed by dividing the size of the intersection of the sets by the size of the union of the sets.

**Table 6** Example sentences of contradiction via negation subset

| Sentence 1 (original sentence) | Sentence 2 (negated sentence) |
| --- | --- |
| Rip1 was reported to interact with rip3. | Rip1 was reported to not interact with rip3. |
| Moreover, other reports have also shown that necroptosis could be induced via modulating rip1 and rip3. | Moreover, other reports have also shown that necroptosis could not be induced via modulating rip1 or rip3. |

Table 7 Example sentences of contradiction via antonyms subset

| Sentence 1 (original sentence) | Sentence 2 (negated sentence) |
|---|---|
| When expressed alone in primary cells however, oncogenic ras induces premature senescence, a putative tumour suppressor mechanism to protect from uncontrolled proliferation. | When expressed alone in primary cells however, oncogenic ras inhibits premature senescence, a putative tumour suppressor mechanism to protect from uncontrolled proliferation. |
| Two recent studies used rnai-mediated tet2 knock-down in vitro to suggest that tet2 depletion led to impaired hematopoietic differentiation and to preferential myeloid commitment. | Two recent studies used rnai-mediated tet2 knock-down in vitro to suggest that tet2 depletion led to enhanced hematopoietic differentiation and to preferential myeloid commitment. |

$$J(A,B) = \frac{|A \cap B|}{|A \cup B|}$$

Applied to the BIOSSES benchmark, set A and B comprise the unique words of sentence 1 and sentence 2, respectively.

Q-gram similarity

A q-gram is a contiguous sequence of q items from a given string, where the items can be letters or words for example. Q-gram similarity is defined as the number of q-grams shared by the respective strings and is calculated by dividing the number of q-gram matches of the second string by the number of possible q-grams determined by the first string.

Vector-based

Cosine similarity

Cosine similarity is a widely used metric for semantic similarity. It measures the similarity of two vectors, in our case sentence vectors, based on the cosine of the angle between them.

$$\cos(\theta) = \frac{\mathbf{A} \cdot \mathbf{B}}{\|\mathbf{A}\|\|\mathbf{B}\|}$$

## Endnotes

[1]https://www.nltk.org/
[2]https://www.ranks.nl/stopwords
[3]http://ftp.ncbi.nlm.nih.gov/pub/pmc
[4]https://www.nltk.org/
[5]https://github.com/klb3713/sentence2vec
[6]https://github.com/tensorflow/models/tree/master/research/skip_thoughts
[7]https://github.com/tmikolov/word2vec
[8]https://github.com/klb3713/sentence2vec

## Additional file

**Additional file 1:** Supplementary tables and figures. (DOCX 181 kb)

## Abbreviations

C-BOW: Continuous-bag-of-words; PMC: PubMed Central; PV-DBOW: Paragraph Vector Distributed Bag of Words; PV-DM: Paragraph Vector Distributed Memory

## Acknowledgements
We want to thank the development teams behind fastText and sent2vec for making these software packages openly available. Furthermore we want to thank Chris Shallue for making his implementation of skip-thoughts available.

## Funding
A part of the research leading to these results has received funding from the European Community's Horizon 2020 Programme under grant agreement No. 668353 (U-PGx). The funding body was not involved in the design of the study and collection, analysis, and interpretation of data or in writing the manuscript.

## Availability of data and materials
Datasets generated and/or analyzed during the current study and Jupyter notebooks are available through Github at https://github.com/kathrinblagec/neural-sentence-embedding-models-for-biomedical-applications

## Authors' contributions
MS devised the study, KB carried out experiments and conducted data analysis. KB and HX conducted data preparation. MS, KB, HX and AA participated in authoring the manuscript. All authors have read and approved the manuscript.

## Ethics approval and consent to participate
Not applicable.

## Consent for publication
Not applicable.

## Competing interests
The authors declare that they have no competing interests.

## Publisher's Note
Springer Nature remains neutral with regard to jurisdictional claims in published maps and institutional affiliations.

Received: 30 July 2018 Accepted: 2 April 2019
Published online: 11 April 2019